\pdfoutput=1

\documentclass[11pt]{article}

\newif\ifarxiv
\arxivtrue

\usepackage[]{acl}

\usepackage{times}
\usepackage{latexsym}

\usepackage[T1]{fontenc}

\usepackage[utf8]{inputenc}

\usepackage{microtype}

\usepackage{color}
\usepackage{lipsum}
\usepackage{colortbl}

\def\parsum#1{\bgroup \textcolor{blue}{Paragraph summary: #1}\egroup}
\def\sectionsum#1{\bgroup \textcolor{green}{Section content: #1}\egroup \\}

\usepackage{tabularx}
\usepackage{tablefootnote}
\usepackage{graphicx}

\usepackage{booktabs}

\usepackage{newtxmath} 

\usepackage{multirow}
\usepackage{subfig} 

\newcommand{\bert}{\textsc{Bert}}
\newcommand{\bertbs}{\textsc{Bert-Base}}
\newcommand{\lbert}{\textsc{Legal-Bert}}
\newcommand{\bertsc}{\textsc{Bert-Sc}}

\title{Privacy-Preserving Models for Legal Natural Language Processing}

\author{Ying Yin \and Ivan Habernal \\
	Trustworthy Human Language Technologies \\
	Department of Computer Science, Technical University of Darmstadt \\
	\texttt{ivan.habernal@tu-darmstadt.de}\\
	\url{www.trusthlt.org}
}

\begin{document}

\ifarxiv
\onecolumn
\noindent \textbf{Privacy-Preserving Models for Legal Natural Language Processing}

\medskip
\noindent Ying Yin, Ivan Habernal

\bigskip
This is a \textbf{camera-ready version} of the article accepted for publication at the \emph{Natural Legal Language Processing Workshop 2022} co-located with EMNLP. The final official version will be published in the ACL Anthology later in 2022: \url{https://aclanthology.org/}

\medskip
Please cite this pre-print version as follows.
\medskip

\begin{verbatim}
@InProceedings{Ying.Habernal.2022.NLLP,
  title     = {{Privacy-Preserving Models for Legal Natural
                Language Processing}},
  author    = {Ying, Yin and Habernal, Ivan},
  booktitle = {Proceedings of the Natural Legal Language Processing
               Workshop 2022},
  pages     = {(to appear)},
  year      = {2022},
  address   = {Abu Dhabi, UAE}
}
\end{verbatim}
\twocolumn

\fi

\maketitle
\begin{abstract}
Pre-training large transformer models with in-domain data improves domain adaptation and helps gain performance on the domain-specific downstream tasks. However, sharing mo\-dels pre-trained on potentially sensitive data is prone to adversarial privacy attacks. In this paper, we asked to which extent we can guarantee privacy of pre-training data and, at the same time, achieve better downstream performance on legal tasks without the need of additional labeled data. We extensively experiment with scalable self-supervised learning of transformer models under the formal paradigm of differential privacy and show that under specific training configurations we can improve downstream performance without sacrifying privacy protection for the in-domain data. Our main contribution is utilizing differential privacy for large-scale pre-training of transformer language models in the legal NLP domain, which, to the best of our knowledge, has not been addressed before.\footnote{\url{https://github.com/trusthlt/privacy-legal-nlp-lm}}
\end{abstract}

\section{Introduction}
\label{sec:introduction}

Transformer-based models \citep{Vaswani.et.al.2017,Devlin.et.al.2019.NAACL} trained in a self-supervised fashion on a huge collection of freely accessible Web texts belong to the currently most successful techniques for almost any downstream NLP task across languages or domains. Their ability to `learn' certain language properties \citep{Rogers.et.al.2020.BERT} and the need of having only a small amount of labeled data in the target domain for fine-tuning makes them superior to other approaches \citep{Brown.et.al.2020.GPT3}. Moreover, additional pre-training with unlabeled target-domain data typically boosts their performance further \citep{Chalkidis.et.al.2020.Findings}.

However, when it comes to preserving private information contained in the original large unlabeled text data, transformer models tend `remember' way too much. \citet{Carlini.et.al.2020.arXiv} show that it is possible to extract verbatim sensitive information from transformer models, such as names and addresses, even when such a piece of information had been `seen' by the model during pre-training \emph{only once}. Current transformer models thus represent a threat to privacy protection, which can have harmful consequences if such models trained on very sensitive data are published, as is the current trend in sharing pre-trained models.

In the legal domain, sensitive information, including names, addresses, dates of birth, are important part of many documents, such as court decisions. Especially in countries with the case-law system, court decisions make the largest fraction of legal texts. However, transformer models pre-trained on such corpora do not protect personal information by design, and ad-hoc solutions, e.g. whitening names in the original texts, are prone to errors and potential reconstruction attacks \citep{Lison.et.al.2021.ACL,Pilan.et.al.2022.arXiv}.

Existing approaches to privacy-preserving deep learning have adapted differential privacy (DP) \citep{Dwork.Roth.2013}, a rigorous mathematical treatment of privacy protection and loss. In particular, stochastic gradient descent with DP (DP-SGD) has been successfully applied to various NLP problems \citep{Senge.et.al.2021.arXiv,Igamberdiev.Habernal.2022.LREC}, including transformer pre-training \citep{Hoory.et.al.2021.EMNLPfindings, Anil.et.al.2021.arXiv}. However, how well DP-regimes perform in the legal domain, pre-trained and fine-tuned across various downstream legal-NLP tasks, remains an open question.

This paper addresses the following three research questions. First, what are the best strategies for pre-training transformer models to be applied in the legal domain? Second, does DP-SGD training scale up to tens of gigabytes of pre-training data without ending up with an extremely big privacy budget?
Finally, can large-scale privacy-preserving transformers compete to their small-scale non-private alternatives?

\section{Related work}
\label{sec:related.work}

\paragraph{Transformer models in legal NLP}

Large contextual LMs based on transformer architecture \citep{Vaswani.et.al.2017} are the state of the art in numerous NLP tasks.
Domain adaptation aims to improve the model performance on downstream tasks in a specialized domain. A common approach is to pre-train \bert\ \citep{Devlin.et.al.2019.NAACL} with a large collection of unlabeled in-domain texts.
In the legal domain, \citet{Chalkidis.et.al.2020.Findings} provide a systematic investigation of possible strategies for \bert\ adaptation and published their model as \lbert. Their work shows that both training \bert\ form scratch or further pre-training the existing general \bertbs\footnote{\bertbs\ stands for `bert-base-uncased' from \url{https://huggingface.co/bert-base-uncased}.} model with legal corpora achieve comparable performance gains. Besides, broader hyper-parameter search has large impact on the downstream performance. \citet{Zheng.et.al.2021.ICAIL} point out that despite the uniqueness of legal language, domain pre-training in the legal field rarely show significant performance gains probably due to the lack of appropriate benchmarks that are difficult enough to benefit from pre-training on law corpora. To address this issue, they release a new benchmark called CaseHOLD that gains up to 6.7\% improvement on macro F1 by additional domain pre-training.
In the legal field, the vast majority of benchmarks exhibit small performance gains after further pre-training \bert\ on law datasets \citep{elwany2019bert,Chalkidis.et.al.2020.Findings}.
However, existing research on legal language models has not considered privacy of the textual datasets.

\paragraph{Privacy-preserving NLP with differential privacy}

Large machine learning models including transformer-based LMs can be prone to privacy attacks such as membership inference attack \citep{shokri2017membership, hayes2019logan, Carlini.et.al.2020.arXiv}, which means it is possible to predict whether or not a data record exists in the model's training dataset given only black-box query access to the model. It hinders the application of such models on numerous real-word tasks involving private user information. To mitigate this limitation, many recent studies devote to privacy-preserving algorithm for large NLP models.

Differential Privacy (DP) \citep{dwork2006calibrating, Dwork.Roth.2013} has been taken as the gold-standard approach to ensure privacy for sensitive dataset. The main goal of privacy-preserving data analysis is to enable meaningful statistical analysis about the database while preventing leakage of individual information. The intuition behind DP is that an individual's data can't be  revealed by a statistical release of the database regardless of whether or not the individual is present in the database, thus any individual shouldn't have significant influence on the statistical release. 
We formally introduce DP in Section~\ref{sec:appendix-dp}.

Unlike works focusing of privatization of individual texts \citep{Habernal.2021.EMNLP,Habernal.2022.ACL,Igamberdiev.et.al.2022.COLING}, applying DP to training neural networks is typically done through differentially-private stochastic gradient descent (DP-SGD) \citep{Abadi.et.al.2016.SIGSAC}; see also \citep{Yu.et.al.2019.SP} for a great explanation. Although DP pre-training of \bert\ has been shown to gain performance on a Medical Entity Extraction task \citep{Hoory.et.al.2021.EMNLPfindings}, how well it performs in the legal area still remains an open question.

\subsection{Off-the-shelf strategies for training with differential privacy}
\label{sec:slow.dp}

DP-SGD training often suffers from big running time overhead that comes from the per-sample gradient clipping. Mainstream DL frameworks such as PyTorch and TensorFlow are designed to produce the reduced gradients over a batch that is sufficient for SGD but are unable to compute the per-sample gradients efficiently. A naive way to achieve this is to compute and clip the gradient of each sample in the batch one by one through a for-loop, which is implemented in PyVacy.\footnote{\url{https://github.com/ChrisWaites/pyvacy}} This approach completely loses parallelism and hence dramatically slows down the training speed. A more advanced method is to derive the per-sample gradient formula and compute it in a vectorized form. Opacus\footnote{\url{https://github.com/pytorch/opacus}} implements this by replacing the matrix multiplication between the back-propagated gradients and the activations from the previous layer in the original PyTorch back-propagation with outer products via einsum function \citep{yousefpour2021opacus}. The activations and back-propagated gradients are captured through forward and backward hooks. A disadvantage of this method is that it cannot currently support all kinds of neural network modules. In addition, it is restricted by quadratic memory consumption  \citep{subramani2021enabling}.

\section{Learning with differential privacy}
\label{sec:appendix-dp}

This section formally introduces differential privacy and can be skipped by readers familiar with that topic.

\subsection{Pure Differential Privacy (\boldmath$\varepsilon$-DP)}
\textbf{Definition of \boldmath$\varepsilon$-DP} Given a real number $\varepsilon > 0$, a randomized mechanism (or algorithm) $\mathcal{M}: D \mapsto R$ satisﬁes $\varepsilon$-DP if for any two neighboring input datasets $d, d’ \in D$ that differs in a single element and for any subset of outputs $S \subseteq R$ it holds that
\begin{equation}\label{math:2.1}
\frac{\Pr[\mathcal{M}(d) \in S]}{\Pr[\mathcal{M}(d') \in S]} \leq \exp(\varepsilon),
\end{equation}
where $\Pr$ stands for the probability distribution taken from the randomness of the mechanism, and $\varepsilon$ refers to the privacy budget.

The value of $\varepsilon$ upper-bounds the amount of influence any individual data has on the mechanism's outputs. Smaller $\varepsilon$ value means stronger privacy guarantee. However, there is no conclusive answer to how small we should set $\varepsilon$ to prevent information leakage in practice. The general consensus is that $\varepsilon \leq 1$ would indicate strong privacy protection, while $\varepsilon \geq 10$ possibly doesn't guarantee much privacy, although the value is application-specific.\footnote{\url{https://programming-dp.com/}}

The above definition implies that the outputs of the mechanism should not differ much, with or without any specific data record. In this case, an adversary can’t infer whether or not a record exists in the input dataset from the outputs of the mechanism, which prevents the extraction of individual training data from a pre-trained model. 

The \emph{sensitivity} of a mechanism $\mathcal{M}$ is the upper bound of the amount of output difference when it's input changes by one entry. Formally, the Global Sensitivity ($GS$) of $\mathcal{M}$ is given by 
\begin{equation}\label{math:2.2}
GS(\mathcal{M})=\max_{d, d': |d|=|d'| \pm 1} |\mathcal{M}(d) - \mathcal{M}(d')|,
\end{equation}
where d and d' are neighboring datasets. The "global" means this holds for any pair of neighboring datasets, as opposed to the "local" sensitivity with one of the datasets fixed. For example, sensitivity of the counting query that computes how many entries in a database is 1.

There are two important properties of DP: Sequential composition and post-processing.
\begin{itemize}
	\item \textbf{Sequential Composition} For mechanisms $\mathcal{M}_1(d)$ satisfies $\varepsilon_1$-DP and $\mathcal{M}_2(d)$ satisfies $\varepsilon_2$-DP, the mechanism $\mathcal{M}(d)=(\mathcal{M}_1(d), \mathcal{M}_2(d))$ that releases both results satisfies $(\varepsilon_1+\varepsilon_2)$-DP.
		
	\item \textbf{Post Processing} If a mechanism $\mathcal{M}(D)$ satisfies $\varepsilon$-DP, then after performing arbitrary function f on $\mathcal{M}(D)$, the mechanism $f(\mathcal{M}(D))$ still satisfies $\varepsilon$-DP.
\end{itemize}

These properties facilitate the design and analysis of a DP algorithm. The composability enables the track of privacy loss for algorithms that traverse the dataset multiple times, and the post processing property ensures that a DP algorithm is robust to privacy attack with auxiliary information. Moreover, advanced composition exists for approximate DP that provides tighter upper bound of privacy.

\subsection{Appropriate Differential Privacy (\boldmath$(\varepsilon, \delta)$-DP)}
\textbf{Definition of \boldmath$(\varepsilon, \delta)$-DP}  Approximate DP relaxes the pure $\varepsilon$-DP requirement by introducing a "failure probability" $\delta$. Similar to the definition of $\varepsilon$-DP, given real numbers $\varepsilon > 0$ and $\delta > 0$, we say a mechanism $\mathcal{M}: D\rightarrow R$ satisfies $(\varepsilon, \delta)$-DP if for all adjacent inputs $d, d’ \in D$ and all $S \subseteq R$, we have 
\begin{equation}\label{math:2.4}
\Pr[\mathcal{M}(d) \in S] \leq e^\varepsilon \Pr[\mathcal{M}(d') \in S] + \delta
\end{equation}
The pure $\varepsilon$-DP is equivalent to $(\varepsilon, 0)$-DP. A non-zero item $\delta$ allows the mechanism fails to be $\varepsilon$-DP with probability $\delta$. This sounds a bit scary, since under certain probability we get no guarantee of privacy at all and there is a risk of compromising the whole dataset. Therefore, the value of $\delta$ must be small enough, preferably less than one over the size of dataset (i.e. $\frac{1}{|D|}$) in order to deliver meaningful results. One of the biggest advantage of $(\varepsilon, \delta)$-DP is that even with negligible $\delta$, it can significantly reduce the sample complexity compared to the pure DP \citep{beimel2014private, steinke2015between, bun2018fingerprinting}. Roughly speaking, given the same size of dataset, $(\varepsilon, \delta)$-DP can achieve higher statistical accuracy than $\varepsilon$-DP while preserving the privacy. Additionally, the $(\varepsilon, \delta)$-DP mechanisms in practice usually don't fail catastrophically and release the whole dataset. Instead, they fail gracefully and still satisfy $c\varepsilon$-DP for some value $c$ in the case of failure probability. For these reasons, approximate DP becomes popular in real applications.

\textbf{The Gaussian Mechanism} A Gaussian mechanism that satisfies $(\varepsilon, \delta)$-DP can be obtained by injecting Gaussian noise as follows
\begin{equation}\label{math:2.5}
\mathcal{M}_{G}(x,f,\varepsilon, \delta)=f(x)+\mathcal{N}(0, \frac{2S^2 \ln(\frac{1.25}{\delta})}{\varepsilon^2}).
\end{equation}

\subsection{Deep Learning with DP} \label{2.3DLwithDP}
In general, the goal of deep learning is to optimize the model parameters so that the output of the loss function is minimized. This optimization is usually achieved by Gradient Descent and its variants. Basically, the model are learned from the gradient of the loss outputs w.r.t. the model parameters. Take the mini-batch Stochastic Gradient Descent (SGD) as example, at each step $t$, a certain number of randomly selected training samples $\{ \boldsymbol{x}_i \ | i \in \boldsymbol{B}_t, \boldsymbol{B}_t \subseteq \{1, ..., N\} \}$\footnote{ N is the total number of training examples.} are fed into the loss function $\mathcal{L}$ and the average of their output gradients are calculated as an estimate of the loss gradient w.r.t the model weights $\boldsymbol{\theta}$, which is then multiplied by the learning rate $\eta$ for Gradient Descent. This can be formulated as follows:
A DP algorithm has certain guarantee that it doesn't leak individual training examples. In the Gradient Descent algorithm, the only access to the training examples is occurred in the computation of the gradient. Therefore, one way to achieve DP is through introducing noise into the gradient before the update of model weights. If the access to the gradient calculated via training data remains DP, then the resulting model is DP according to the post-processing property. Based on this, \citet{Abadi.et.al.2016.SIGSAC} propose a sophisticated method that turns the mini-batch SGD algorithm into DP, named DP-SGD, which has become a dominant approach to privacy-preserving deep learning. 

DP-SGD primarily modifies two places of the original SGD algorithm to ensure DP. One is to clip the per-example gradients so that the Euclidean-norm (L2-norm) of each individual gradient does not exceed a pre-defined upper bound $C$, which corresponds to a constraint for the sensitivity of gradient. The other one is to add scale-specific Gaussian noise $\mathcal{N}$ into the aggregated clipping gradient:
\begin{equation}
\begin{aligned}
	\boldsymbol{\theta}_t \gets \boldsymbol{\theta}_{t-1} - \frac{\eta}{|\boldsymbol{B}_t|} \left( 
	\sum_{\forall{i} \in \boldsymbol{B}_t} \right.
	&\mathrm{clip}
	(\triangledown_{\boldsymbol{\theta}_{t-1}}\mathcal{L}(\boldsymbol{\theta}_{t-1}, \boldsymbol{x}_i), C) \\
	&+ \left. \mathcal{N}(0, \sigma^{2}C^{2}\boldsymbol{I})
	\right),
\end{aligned}
\end{equation}
where $\sigma$ refers to the a constant called "noise multiplier", higher $\sigma$ produce stronger privacy guarantee. According to the definition of \ref{math:2.5}, the modified SGD is a Gaussian mechanism that satisfies $(\varepsilon, \delta)$-DP. The choice of Gaussian noise is due to the high-dimensionality of the gradient. L2-norm can be applied to measure the sensitivity of a high-dimensional vector-valued function for Gaussian mechanism, which yields much lower sensitivity than Laplace mechanism that only allows the use of L1-Norm, thus much less noise needs to be added to the gradient. Moreover, \citet{Abadi.et.al.2016.SIGSAC} introduce the Moments Accountant for tighter estimation of the privacy cost. Despite its simplicity, DP-SGD brings successes in many deep learning fields.

\section{Experimental setup and data}
\label{sec:experiments.data}

Our experiments aim to find a strategy where \bert\ can benefit from additional domain-specific DP pre-training. Moreover, we explore the trade-off between the privacy budget and model utility under the best setup we obtain.

\paragraph{Privacy-protecting scenario}
In our scenario, we assume that we publish a pre-trained or fine-tuned model, to which an adversary has a full access \citep{Yu.et.al.2019.SP}. The model can be pre-trained on (a) a public general dataset and (b) in-domain, potentially sensitive legal documents, and fine-tuned on (c) a public down-stream task. Our aim is to protect (b) from the adversary.

\subsection{Pre-training \bert\ from scratch} \label{4.1PT}
\bert\ pre-training is a very expensive task, especially with DP. While further pre-training the existing \bertbs\ can take advantage of the already learned language features and greatly reduce the convergence time, the original generic vocabulary remains unchanged. A generic vocabulary might not match the specialized legal terminology and could lead to drastic splitting into sub-word units and reducing semantic expressiveness \citep{Zheng.et.al.2021.ICAIL,Habernal.et.al.2022.AILaw.arXiv}. To address this problem, pre-train \bert\ from scratch with a custom legal tokenizer built on the legal corpus using the WordPiece algorithm \citep{wu2016google}.

In order to investigate the effect of domain vocabulary on model performance and also follow the setup in \citet{Hoory.et.al.2021.EMNLPfindings} that successfully introduce DP to the pre-training of medical \bert\, our pre-training from scratch can be roughly divided into three steps:
\begin{enumerate}
	\item  Generating a domain-specific tokenizer and vocabulary set based on the legal corpus.\footnote{Here we use \texttt{BertWordPieceTokenizer} from \url{https://github.com/huggingface/tokenizers}, we set the vocabulary size to 30,522 which is the same as with \bertbs.}
	\item Pre-training \bert\ from scratch on the generic BookCorpus and Wikipedia dataset using the domain-specific tokenizer.
	\item Further pre-training \bert\ with DP on the legal corpus.
\end{enumerate}
In spite of that the first step also involves access to the legal corpus and may cause information leakage, there is no good solution to convert the WordPiece algorithm into DP with tight privacy bound. We leave this problem to future work. Currently, we only ensure privacy during the pre-training on the legal corpus. The second step only uses the general corpora and thereby has no privacy issue. We don't use the legal corpora at the beginning of the pre-training because the overhead to train DP \bert\ from scratch is too expensive. We call the model trained with the first two steps \bertsc.

\subsection{Further pre-training \bertbs} \label{4.1.1FTsetup}
Continuing the pre-training of \bertbs\ with legal-domain corpora is an economical and effective way for domain transfer. We start with a small-scale pre-experimental corpus to quickly investigate the effectiveness of additional domain pre-training with different hyper-parameter settings. Afterward, we scale up the training on the full legal corpus and focus on the batch size and learning rate tuning. In order to avoid overfitting, 5\% of the pre-training data is kept as a validation set, on which the sum loss of the MLM (masked language modeling) and NSP (next sentence prediction) objectives and their accuracy is evaluated at each checkpoint.

\section{Downstream tasks and datasets}

We experiment with two downstream benchmark datasets, Overrruling and CaseHOLD, on which we fine-tune our pre-trained models.\footnote{Hyperparameters for the downstream tasks are discussed in Appendix~\ref{app:hyperparam-downstream}.} Note that for the downstream tasks, we do not use differential private training.

The Overruling dataset \citep{Zheng.et.al.2021.ICAIL} corresponds to a binary classification task that predicts if a sentence has the meaning of voiding a legal decision made in a previous case, which is important to ensure the correctness and validity of legal agreements. The sentences in the dataset are sampled from the Casetext law corpus, where positive overruling examples are manually annotated by lawyers, and negative examples are automatically generated by randomly sampling the Casetext sentences because over 99\% of them are non-overruling. The complete dataset contains 2400 items and the two classes are balanced. It is a relatively simple task that has already achieved state-of-the-art performance on \bertbs\ model, since the positive examples explicitly contain `overrule' or words with similar meaning such as disapprove, decline, reject, etc., which makes them highly distinguishable from the negative ones. 

The CaseHOLD (Case Holdings on Legal Decisions) is a multiple-choice QA task to select a correct holding statement among 5 potential answers that matches the given citing context from a judicial decision. \citet{Zheng.et.al.2021.ICAIL} construct the dataset by extracting the legal citations and the accompanying holding statements from the corpus of U.S. CaseLaw and using them as questions and answers respectively. Here the cases contained in the CaseHOlD are removed from our legal pre-training corpus according to the case IDs they provide. Moreover, they search for propositions that are semantically similar to the corresponding answer from other extracted holding statements as the wrong answers according to the TF-IDF similarity between them, which makes the CaseHOLD a multiple-choice QA task. The labels of the correct answers are uniformly distributed within the 5 indices 0-4. Excluding some samples containing invalid labels, the full dataset we use has a total of 52,978 items. It is a challenging task and yields only a macro F1 of around 0.613 using the general \bertbs\ \citep{Zheng.et.al.2021.ICAIL}. We use it to investigate whether a sufficiently difficult legal task benefits from additional domain pre-training in the private preserving scenario.

\section{Our approach to pre-training legal transformer models with DP}

\subsection{Datasets for pre-training}
For the in-domain pre-training with differential privacy, we prepare a legal corpus consisting of 14GB legal texts that are collected from three different resources (see Table \ref{table:3.1}). Although these are public datasets, we treat them as if they were private, containing sensitive data whose leakage from the pre-trained models should be prevented. For compiling and caching the large-scale pre-training corpora, we leverage the HuggingFace Datasets library\footnote{\url{https://huggingface.co/docs/datasets/}} based on Arrow, which allows fast lookup for big datasets by building a memory-mapped cache on disk. 

\begin{table}[h!]
	\centering
	\begin{tabular}{ lrr } 
		\toprule
		Source & Documents & Size (GB) \\ 
		\midrule
		Sigma Law\tablefootnote{\url{https://osf.io/qvg8s/}} & 39,155 & 1.2 \\ 
		LEDGAR\tablefootnote{\citet{Tuggener.et.al.2020.LREC}} & $\approx$ 300,000 & 0.2 \\ 
		Case Law\tablefootnote{\url{https://case.law}}  & $\approx$ 28,300,000 & 12.6 \\ 
		\bottomrule
	\end{tabular}
	\caption{Details of the legal corpora for pre-training.}
	\label{table:3.1}
\end{table}

\subsection{Scalable pre-training with DP}

In section~\ref{sec:slow.dp} we discussed the shortcomings of off-the-shelf DP-SGD implementations in mainstream frameworks. We carried out preliminary experiments and found that these shortcomings make DP-SGD pre-training infeasible due to 12 to 28-times longer runtime per epoch.

The training speed of DP-SGD can be significantly improved by vectorization, just-in-time (JIT) compilation and static graph optimization using JAX framework,\footnote{\url{https://github.com/google/jax}} which is defined by JIT compilation and automatic differentiation built up on the XLA compiler \citep{subramani2021enabling}. The core transformation methods of main interest in JAX includes \verb|grad|, \verb|vmap|, \verb|jit|, and it allows us to arbitrarily compose these operations.
In the DP-SGD scenario, \verb|grad| can automatically compute the gradients of the loss objective w.r.t. the model parameters, and combing \verb|vmap| enables efficient computation of per-example gradients by vectorizing the gradient calculation along the batch dimension.

Furthermore, the DP-SGD step can be decorated by \verb|jit| to leverage XLA compiler that has proven acceleration in \bert\ MLPerf submission. Although JAX shows great advantages over other mainstream DP frameworks and libraries on a wide variety of networks such as Convolutional Neural Network (CNN) and Long-Short Term Memory network (LSTM) in \citet{subramani2021enabling}, how much speedup it can produce on large transformer-based LMs remains unknown.

To investigate this, we implementat a JAX version of DP \bert\ based on FlaxBert models,\footnote{\url{https://huggingface.co/docs/transformers/model_doc/bert}} which provides transformers with JAX/FLAX backend including \bert. We adapt its training step into DP by adding the per-sample gradients clipping before aggregation and introducing randomly sampled Gaussian noise to the reduced gradients. Moreover, we use the strategy of gradient accumulation to enable DP pre-training with arbitrarily large batch sizes. Specifically, a training step is split into many iterations such that each iteration handles a shard of examples that the GPU\footnote{All the experiments are carried out on an NVIDIA A100 40GB.} memory can maximally hold, and the clipped per-example gradients are accumulated over iterations within a batch.

\subsubsection{Finding optimal hyper-parameters}

Our starting point to further pre-training with differential privacy is the uncased \bertbs\ model that contains 110M parameters. For the optimization, we use Adam with weight decay (AdamW, \cite{loshchilov2018fixing}) and a linear learning rate schedule, which consists of a warm-up phrase followed by a linear decay. The warm-up steps are set to roughly 5\% of the total training steps with a lower bound of 25. In addition, we use the TensorFlow privacy library\footnote{\url{https://github.com/tensorflow/privacy}} based on Rényi DP (RDP) \citep{mironov2017renyi, mironov2019r} for the track of privacy, which can be converted to a standard $(\varepsilon, \delta)$-DP but provides a tighter composition for Gaussian mechanism than directly using $(\varepsilon, \delta)$-DP. The method takes the noise multiplier $\sigma$ as input and calculates the privacy budget $\varepsilon$ for each step. Conversely, we obtain the desired $\varepsilon$ by the binary search for an optimal noise multiplier that leads to a privacy budget close enough to the target $\varepsilon$ in a proper range.  

In our experiments, the gradient clipping norm and the weight decay are less significant factors, and we fix them to 1.0 and 0.5 accordingly. To study the influence of the batch size, we keep the privacy $\varepsilon$ to 5, which is considered as a sweet point between a very strong privacy guarantee 1 and a weak guarantee 10. In order to avoid overfitting, 5\% of the pre-training data is kept as a validation set, on which the sum loss of the MLM and NSP objectives and their accuracy is evaluated at each checkpoint.

\section{Results and analysis}
\label{sec:results.analysis}

\textbf{Baselines} Our baseline results (Table \ref{table:5.1}) are reported from \bertbs\ and \bertsc\ with tuned hyper-parameters with no privacy gurantees. \bert\ trained from scratch with a custom legal vocabulary (\bertsc) slightly outperforms vanilla \bertbs.

\begin{table}
	\centering
	\begin{tabular}{lrr}
		\toprule
		Model & Overruling & CaseHOLD \\ 
		\midrule
		\bertbs\ & 0.971 & 0.617 \\

		\bertsc\ & 0.975 & 0.618 \\ 
		\bottomrule
	\end{tabular}
	\caption{Baseline Macro-$F_1$ scores without any domain pre-training.}
	\label{table:5.1}
\end{table}

\paragraph{Small-scale pre-training with DP}

We experimented with further pre-training of two baseline models on a small-scale 2.3GB  legal sub-corpus. The goal was to efficiently explore the effect of several key hyper-parameters on DP training with a small amount of data. We trained for 29k steps at batch size 256.

\begin{table}[]
	\centering
	\begin{tabular}{rrrr}
		\toprule
		$\sigma$ & $\varepsilon$ & Overruling & CaseHOLD \\ 
		\midrule
		\multicolumn{4}{l}{\bertbs} \\
		-- & -- &  0.975 & 0.652 \\
		1e-5 & $\infty$ &  0.967 & 0.648 \\ 
		0.1 & 4e+5 &  0.971 & 0.616 \\
		\vspace{1em}0.5 & 3.726 &  0.969 & 0.613 \\
		
		\multicolumn{4}{l}{\bertsc} \\
		-- & -- &  0.969 & 0.647 \\
		1e-5 & $\infty$ &  0.967 & 0.645 \\ 
		0.1 & 4e+5 &  0.967 & 0.618 \\
		0.5 & 3.726 &  0.964 & 0.616 \\
		\hline
	\end{tabular}
	\caption{\label{table:5.2}
		Macro-$F_1$ scores for further small-scale pre-training of \bertbs\ and \bertsc. $\sigma$="--" corresponds to the training without DP.}
\end{table}

Table~\ref{table:5.2} shows that
while both baseline models after further pre-training without privacy achieve $\sim$ 3\% substantial performance gains on CaseHOLD, the results of DP pre-training is disappointing. The benefits of domain training for CaseHOLD task seem to disappear after adding even a small amount of noise ($\sigma =0.1$). The results from $\sigma=0.1$ and $\sigma=0.5$ don't outperform the baseline or are even marginally worse than it. In addition, the legal tokenizer doesn't indicate an advantage over the general one. We conclude that small-scale DP pre-training barely brings any improvement and even hurts the performance. We decide to scale up the training and explore larger batch sizes.

\subsection{Large-scale domain pre-training with DP} \label{5.2large}

As the batch size is one of the most important parameter in DP training, we fix the target privacy budget $\varepsilon$ as 5 and further pre-train \bertbs\ on the large-scale full legal corpus starting with the default parameters (see Table~\ref{table:A.1} in the Appendix). Then we explore the parameter space by gradually increasing the batch size up to $\sim$ 1M and roughly tune the learning rate at the same time. Although we have significantly accelerated the DP training by JAX framework, large-scale DP pre-training is still quite expensive. Due to resource and time constraints, we do not perform a complete grid search but only experiment with the likely best learning rates at each batch size in our experience. 

\begin{figure*}[h!]
	\centering
	\includegraphics[width=0.48\linewidth]{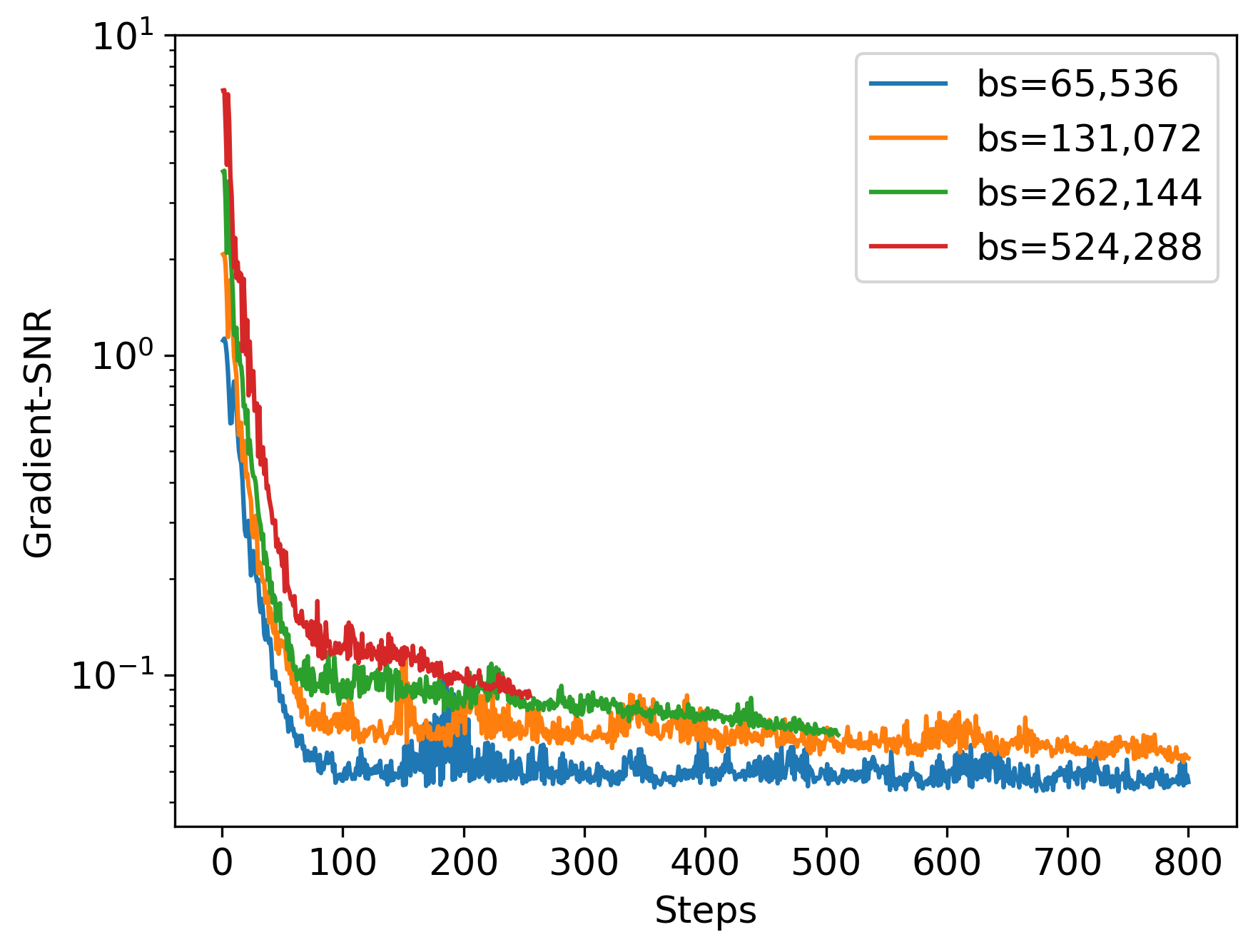} 
	\includegraphics[width=0.48\linewidth]{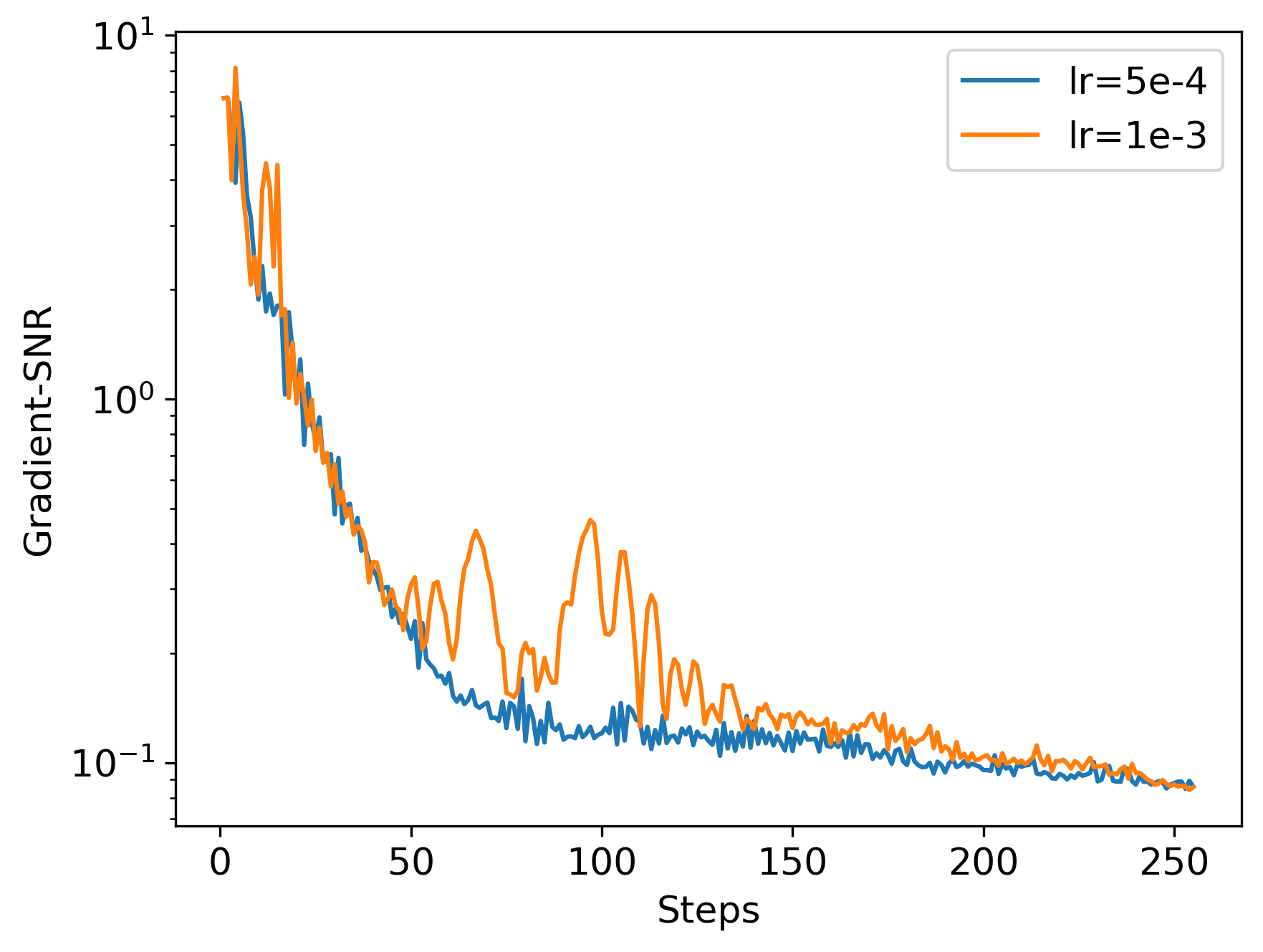}
	\caption{Gradient-SNR over steps for DP pre-training with same privacy budget and fixed epochs while varying the batch size (bs) or learning rate (lr). The left plot shows the trends of SNR at four different batch sizes. For smaller batch sizes, the SNRs after 800 steps are not presented, but they've basically converged to a small value as seen in the figure. Their initial learning rates are uniformly set as 5e-4. The right-hand figure draws the changes on SNR at batch size 524,288 using two different learning rates. }
	\label{fig:5.2}
\end{figure*}

\paragraph{Gradient-SNR} Following the work in \citet{Anil.et.al.2021.arXiv}, we keep track of the gradient signal-to-noise ratio at each step during the DP pre-training of \bert.
Figure \ref{fig:5.2} shows the impact of batch size and learning rate on the Gradient-SNR. In general, the SNR decreases with training and eventually converges to a small value. This is probably due to the fact that the magnitude of the gradient decreases constantly during the learning, whereas the magnitude of the noise remains basically the same, so the ratio of the two keeps shrinking until the gradients become stable. From the left subplot \ref{fig:5.2}(a) we can see that a larger batch size leads to higher Gradient-SNRs. Moreover, the right subfigure (b) shows that an appropriate learning rate can also improve the Gradient-SNR for a certain batch size. However, a too large learning rate leads to dramatic oscillation of Gradient SNR, and the model may move away from the local optima and thus increase the training loss.

\begin{figure*}[t]
	\centering
	\includegraphics[width=\linewidth]{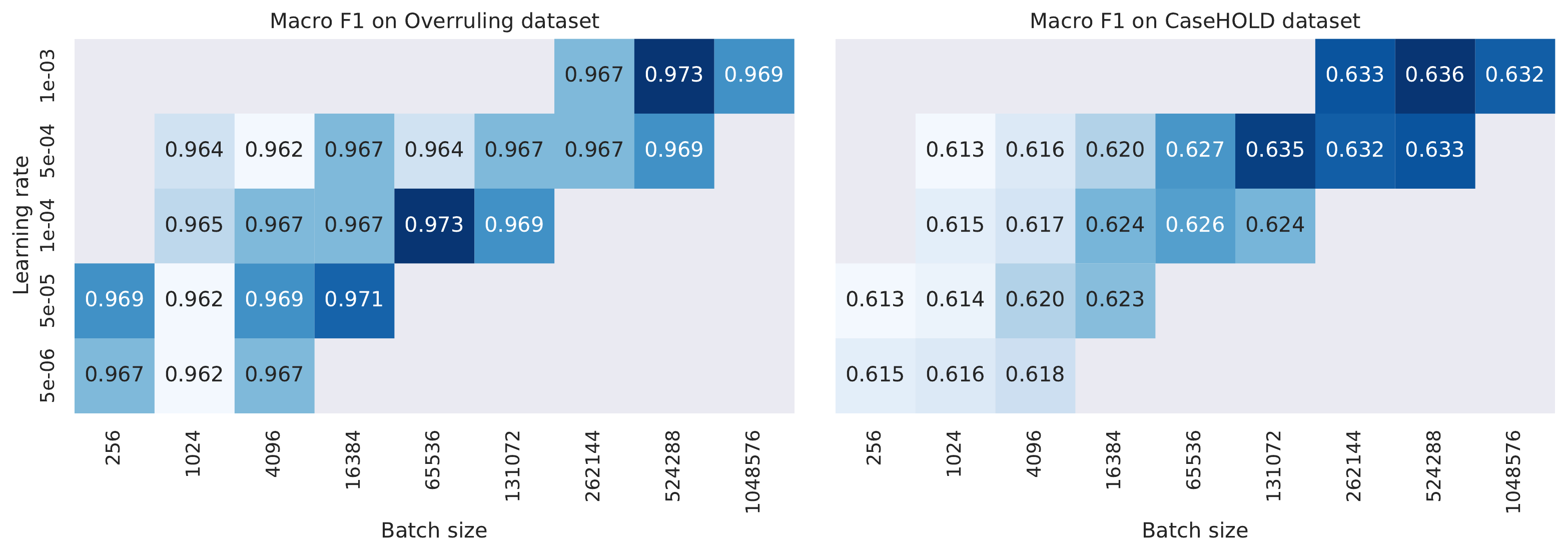}
	\caption{Downstream results obtained by tuning the batch size and learning rate of large-scale domain pre-training when fixing both the target privacy $\varepsilon$ and training epochs to 5. }
	\label{fig:5.3}
\end{figure*}

\paragraph{Results on downstream tasks}

In our experiments with fixed training epochs, the batch size and learning rate jointly influence the performance on CaseHOLD. As can be seen from the bottom subplot of Figure \ref{fig:5.3}, training with unusually large batch sizes and high learning rates (upper right area) produces significantly better Macro F1 scores than using small batch sizes and low learning rates (bottom left area). By scaling up the batch size and tuning the learning rate accordingly, we achieve the best Macro F1 of 0.636 at batch size 524,288 and learning rate 1e-3. This \textbf{outperforms the baseline by almost 2\%}.

As a summary, for a fixed training epoch setting, enlarging the batch size is not always beneficial and tuning the learning rate is crucial as well. However, according to our experiments, DP pre-training of \bert\ with a regular small batch size performs overall very poorly, and it starts to make performance gains on CaseHOLD when the batch size is stepped up to 4,096. We obtain a significant boost when increasing the batch size to 130K+. We conclude that scaling up the batch size and in-domain corpus is necessary to obtain good performance for DP pre-training of BERT in the legal field.

\section{Discussion}

Here we clarify some questions and comments raised by the reviewers.

\paragraph{Is the 2\% improvement worth the effort?} 

We believe so. Let's put our result into a broader context by having a closer look at results achieved by \lbert\ \citep{Chalkidis.et.al.2020.Findings}. On three downstream tasks, they gained similar improvements. First, \emph{``in ECHR-CASES, we [...] observe small differences [...] in the performance on the binary classification task (0.8\% improvement)."} Second, on NER they observed an \emph{``increase in F1 on the contract header (1.8\%) and dispute resolution (1.6\%) subsets. In the lease details subset, we also observe an improvement (1.1\%)."} Finally, on EURLEX57k, they observed \emph{``a more substantial improvement in the more difficult multi-label task (2.5\%) indicating that the LEGAL-BERT variations benefit from in-domain knowledge."} Moreover, our approach achieves similar gains under differential privacy guarantees.

\paragraph{How expensive is DP training?}

\begin{figure}
	\centering
	\includegraphics[width=0.9\linewidth]{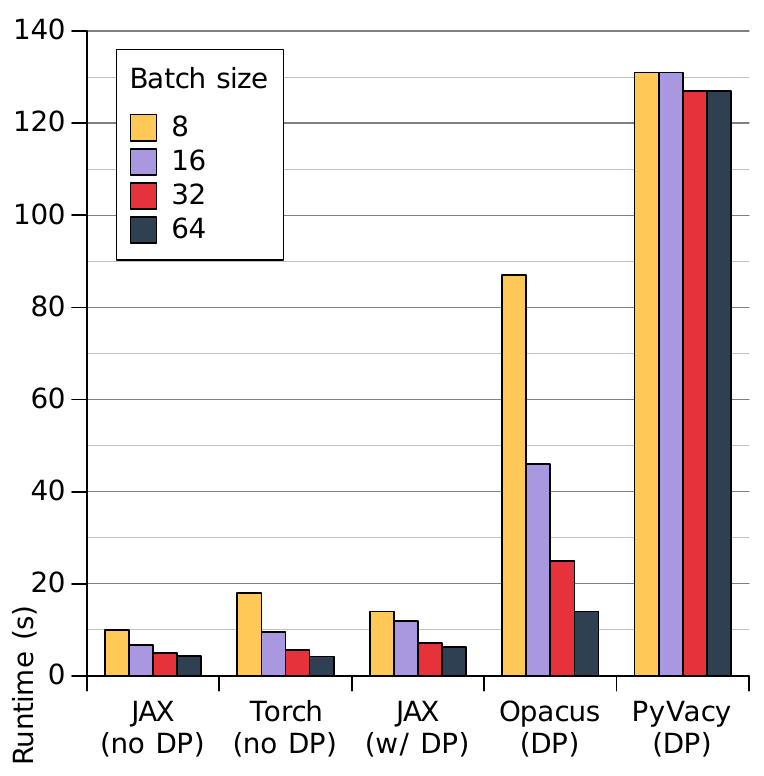}
	\caption{\label{fig:runtime} Runtimes (in seconds) per epoch for fine-tuning BERT on the Overruling binary classification task with different batch sizes and frameworks.}
\end{figure}

We experimentally evaluate the running performance of different frameworks on a binary classification task (Overruling) in both private and non-private cases. Figure \ref{fig:runtime} show the runtimes per epoch taken from the median over 20 epochs of training. In our experiments, Opacus is unable to support BERT's Embedding layer, although we use its official tutorial for training. This also prevents us to use it for the DP pre-training. We freeze its Embedding layer for the fine-tuning, which reduces nearly 22\% training parameters compared to other methods. By doubling the batch size each time, 64 is the maximum batch size that the current GPU can support for JAX framework. Opacus uses a \texttt{BatchMemoryManager} to eliminate the limit of batch size similar to gradient accumulation, but the physical batch size it can achieve is actually much smaller than 64.
This indicates that JAX has higher memory-efficiency than Opacus.
The runtime of all the methods decreases significantly as the batch size grows except for PyVacy.
In summary, due to the performance of JAX in the DP training, the `extra costs' are negligible and allows us to upscale DP pre-training.

\paragraph{The title is misleading, the authors do not propose a privacy-preserving legal NLP model.}

If we take the definition of privacy through the lenses of differential privacy, then our pre-trained model is privacy-preserving; see, e.g., \citet{Yu.et.al.2019.SP} for a terminology clarification, or parallel works with the T5 language model \citep{Ponomareva.et.al.2022.FindingsACL}.

\paragraph{Why even do this?}

The scenario in which we want to protect privacy is the following. Say a company has huge amounts of in-house sensitive legal texts (e.g., contracts) which are valuable for pre-training a LM. This model is likely to be better performing on similar domains, so the company wants to offer an API or provide the model to other parties for further fine-tuning. Without DP, privacy of the pre-training data can be compromised \citep{Pan.et.al.2020.SP,Carlini.et.al.2020.arXiv,Yu.et.al.2019.SP}.

\section{Conclusion}
\label{sec:conclusion}

This paper shows that we can combine large-scale in-domain pretraining for a better downstream performance while protecting privacy of the entire pre-training corpus using formal guarantees of differential privacy. In particular, we implemented highly-scalable training of the \bert\ model with differentially-private stochastic gradient descent and pre-trained the model on $\approx$ 13 GB legal texts, using a decent $\varepsilon = 5$ privacy budget. The downstream results on the CaseHOLD benchmark show up to 2\% improvements over baseline models with tuned hyper-parameters and models trained from scratch with a custom legal vocabulary. Our main contribution is utilizing differentially-private large-scale pre-training in the legal NLP domain. We believe that adapting formal privacy guarantees for training models might help overcome the difficulties of using large but potentially sensitive datasets in the legal domain.

\section*{Acknowledgements}

The independent research group TrustHLT is supported by the Hessian Ministry of Higher Education, Research, Science and the Arts. We thank the anonymous reviewers for their useful feedback.

\bibliography{bibliography}

\appendix

\begin{table*}[t!]
	\centering
	\begin{tabular}{llll}
		\hline
		& Learning Rate & Batch Size & Epochs \\ 
		\hline
		\citet{Devlin.et.al.2019.NAACL} & 2e-5, 3e-5, 4e-5, 5e-5 & 16, 32 & 3, 4 \\
		\hline
		First round & 5e-6, 1e-5, 5e-5, 1e-4 & 8, 16, 32, 64, 128 & max 10, early stop \\ 
		\hline
		Second round & 7e-6, 2e-5, 3e-5, 7e-5 & {16 Overruling; 128 CaseHOLD} & max 10,  early stop \\
		\hline
		Final setup & 1e-5, 3e-5, 5e-5, 7e-5 & {16 Overruling; 128 CaseHOLD} & max 5, early stop \\
		\hline
	\end{tabular}
	\caption{Summary of the hyper-parameter search}
	\label{table:A.1}
\end{table*}

\begin{table*}[t!]
	\centering
	\begin{tabular}{cccccc}
		\hline
		$\omega$ &  FP eval loss & MLM acc & NSP acc & F1 on Overruling & F1 on CaseHOLD \\ 
		\hline
		0.1 & 1.706 & 0.682 &	0.947 & 0.973 & 0.636 \\
		0.5 & 1.701 & 0.681 &	0.947 & 0.973 & 0.636 \\ 
		1.0 & 1.695 & 0.681 &	0.948 & 0.969 & 0.636 \\
		\hline
	\end{tabular}
	\caption{Evaluation results for tuning the weight decay $\omega$ on the best setup (bs=524,288, lr=1e-3).}
	\label{table:5.3}
\end{table*}

\section{Hyperparemeters for downstream tasks}
\label{app:hyperparam-downstream}

As shown by \citet{Chalkidis.et.al.2020.Findings}  and \citet{Zheng.et.al.2021.ICAIL}, downstream hyper-parameters have a significant impact on the evaluation results and an enriched search range is necessary for the legal benchmarks. Therefore, instead of blindly following the recommended search range given by \citet{Devlin.et.al.2019.NAACL}, we perform a broader search through two rounds of coarse- to fine-grained grid search. The details of the searched hyper-parameters are shown in Table \ref{table:A.1}. In the final setup, we fix the batch size as 16 for Overruling and 128 for CaseHOLD, and train for a maximum of 5 epochs. Furthermore, the downstream performances are relatively sensitive to the learning rate, we do a search over \{1e-5, 3e-5, 5e-5, 7e-5\} and the best macro-f1 scores are reported for each pre-trained model.

\section{Additional experiments with limited impact}

\subsection{Weight Decay \boldmath$\omega$} BERT uses layer normalization \citep{ba2016layernorm} that makes the output of a layer independent of the scale of its weights. As explained in \citet{Anil.et.al.2021.arXiv}, the Frobenius norm of the layer weights tends to grow due to the noise introduced in the DP training, which reduces the norm of the gradients and thereby slows down the learning process under the layer normalization. To address this problem, they suggest using a much larger weight decay for Adam optimizer compared to the non-private training. Therefore, we experiment with several different weight decays on the best setup of batch size and learning rate. The results are outlined in Table \ref{table:5.3}. Different from the results in \citet{Anil.et.al.2021.arXiv}, changing the weight decay causes almost no impact on the downstream performance and accuracy of MLM and NSP. One can only observe a negligible decline in loss as the weight decay increases. This is probably because our training starts from a well pre-trained base model, the weight update is more stable than training from scratch.

\subsection{L2 Clipping Norm \boldmath$C$} Recall that the two critical steps in DP-SGD are to clip the L2 norms of per-example gradients to $C$ and to introduce randomly sampled Gaussian noise with standard deviation $\sigma C$. Both steps involve the clipping norm $C$, thus it is likely to be an important hyper-parameter for DP training. We experiment with different values of $C$ in \{0.01, 0.1, 1.0, 10\} at batch size 1024 and $\sigma$ 0.5. However, the MLM and NSP accuracy and downstream performance are almost unchanged when we drastically vary $C$. Hence, we consider that the L2 clipping norm may not be a key factor to DP pre-training and fix it to 1.0 in future experiments based on the common best results of two end tasks.

\end{document}